\documentclass[journal,comsoc]{IEEEtran}
\usepackage{multirow}
\usepackage{amsmath}
\usepackage{booktabs}
\usepackage[T1]{fontenc}
\usepackage{cleveref}

\usepackage{cite}

\ifCLASSINFOpdf
  \usepackage[pdftex]{graphicx}

\else
   \usepackage[dvips]{graphicx}
\fi
\usepackage{amsmath}
\usepackage[cmintegrals]{newtxmath}
\begin{document}
%
\title{Building Emotional Machines: Recognizing Image Emotions through Deep Neural Networks}
%
%
%

\author{Hye-Rin Kim,
        Yeong-Seok Kim,
        Seon Joo Kim,
        In-Kwon Lee 
}

\maketitle

\begin{abstract}
An image is a very effective tool for conveying emotions. 
Many researchers have investigated in computing the image emotions by using various features extracted from images. 
In this paper, we focus on two high level features, the object and the background, and assume that the semantic information of images is a good cue for predicting emotion.
An object is one of the most important elements that define an image, and we find out through experiments that there is a high correlation between the object and the emotion in images.
Even with the same object, there may be slight difference in emotion due to different backgrounds, and we use the semantic information of the background to improve the prediction performance.
By combining the different levels of features, we build an emotion based feed forward deep neural network which produces the emotion values of a given image. 
The output emotion values in our framework are continuous values in the 2-dimensional space (Valence and Arousal), which are more effective than using a few number of emotion categories in describing emotions.
Experiments confirm the effectiveness of our network in predicting the emotion of images.
\end{abstract}


%
\IEEEpeerreviewmaketitle

\section{Introduction}

\begin{figure}
\centering
\includegraphics[width=3.4in]{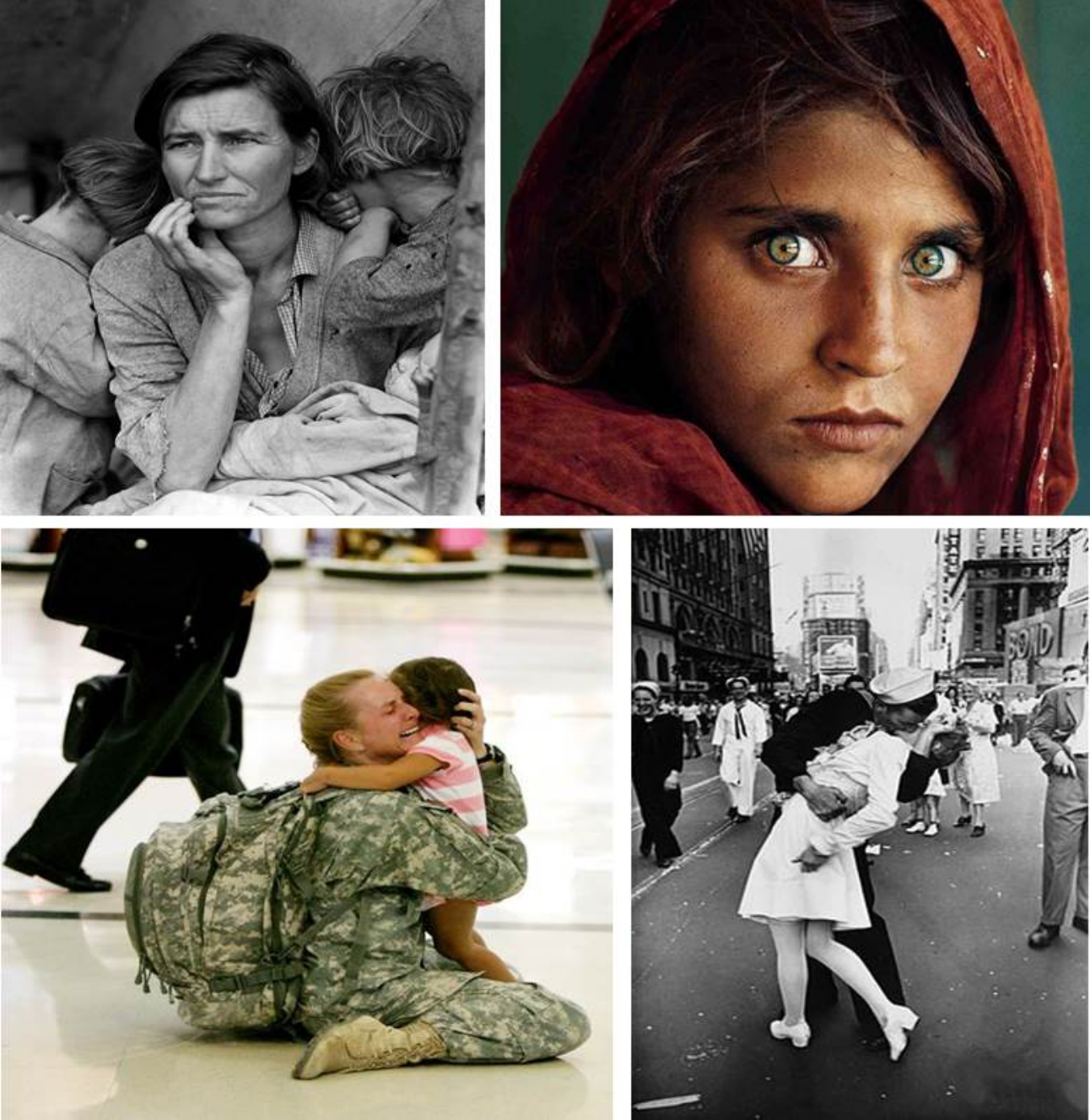}
\caption{Images with different emotions.}
\label{fig:main}
\end{figure}

Images are very powerful tools for conveying moods and emotions as shown in Figure~\ref{fig:main}. 
Through images, people can express their feelings and communicate with other people. 
With the recent development in the deep learning technology, computers have become better at recognizing objects, faces, and actions. 
Computers have also started to write image captions and answer questions about images. But how about emotions? 
Can we teach computers to have similar feeling as humans do when looking at images? 
Predicting evoked emotion from an image is a difficult task and is still in its early stage.

\begin{figure}[t]
\center
\includegraphics[width=1.0\linewidth]{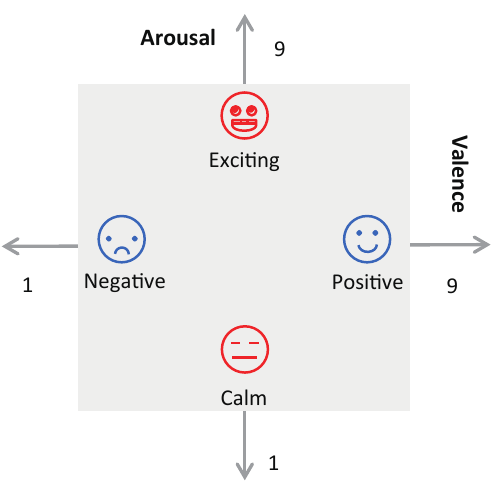}
   \caption{Two dimensional emotion models (Valence and Arousal).}
\label{fig:avmodel}
\end{figure}

As the deep learning technology shows remarkable performance in various computer vision tasks such as image classifcation~\cite{krizhevsky2012imagenet,simonyan2014very,szegedy2015going, he2015deep}, segmentation~\cite{long2015fully,girshick2014rich} and image processing\cite{cheng2015deep,dong2014learning}, several studies have been introduced recently that apply the deep learning for the emotion prediction~\cite{campos2015diving,peng2015mixed,you2015robust,you2016building,xu2014visual}.
Those works mostly use the convolutional neural network (CNN)~\cite{lecun1998gradient}, which has shown better prediction results for the emotion classification compared to the model that uses a shallow network such as the linear model.

The CNN has had a big impact especially in the image classification, and it is an effective network model for learning filters that capture the shapes that repeatedly appear in images.
However, we argue that the learning process for the image emotion prediction should be different from that of image classification.
This is because some images with different appearances can have same emotions, and some images with similar appearances can have different emotions.
For example, an image that includes a person riding a bike and an image that includes a person surfing in the ocean may give the same feelings, even though they look different. 
From this point of view, the performance may be limited if CNN is applied to an emotion prediction system.

In addition, one of the main issues in the emotion recognition is the affective gap.
The affective gap is the lack of coincidence between the measurable signal properties, commonly referred to as features, and the expected affective state in which the user is brought by perceiving the signal~\cite{hanjalic2006emotion}.
To narrow this affective gap, several works proposed emotion classification systems based on the psychology and art theory based high level features such as harmony, movement, rule of third, etc.~\cite{lu2012shape,machajdik2010affective,zhao2014exploring}. 
While those features help to improve the emotion recognition, a better set of features are still necessary.
Distinguishing the effective features among various features is also important. 
As another example, similar to the example above, images with the objects such as guns or sharks arouse scary feeling, while images with babies or flowers lead to more happiness. 
It can be speculated that certain objects will affect the determination of emotions.
Based on this observation, we assume that the main object appearing in the image plays an important role in determining the emotion. The idea is that object categories can be good cues for emotions. 
Through experiments, we show that objects appearing in images are related to emotion values, and objects are used as one of the features of our model. 
Besides, the emotion, even if images include the same object, can vary depending on the background.
We also use the semantic information of the background as our features to improve prediction performance.

Predicting emotions from images is a complex task that is quite different from the object detection or the image classification.
In this paper, we combine high-level features such as the object and the background information extracted from a pre-trained deep network model for the image classification and the segmentation with low-level features such as color statistics to obtain a set of features for the emotion prediction. 
Using this feature set, we design a feed forward deep network (FFNN)~\cite{haykin1999neural} which produces the emotion value for a given image. 

In previous works for the emotion prediction, the emotions are categorized by a few number of classes, such as happy, awe, sad, fear, etc. 
In comparison, we opt for the dimensional model for expressing the emotions~\cite{osgood1952nature}, which is widely used in the field of psychology. 
Specifically, the dimensional model consists of two parameters, Valence and Arousal (Figure~\ref{fig:avmodel}). 
Valence represents the pleasure through the scale from 1 (negative) to 9 (positive). 
Arousal is the level of excitement, which also ranges from 1 (calm) to 9 (excited). Using this model, any emotion can be represented by using these two values in 2D space. As this dimensional model can express more emotions compared to using a discrete set of emotion categories, we train our emotion prediction system based on the V-A model.

The contributions of our work are summarized as follows:
\begin{itemize}
\item We propose an image emotion recognition system which outputs valence-arousal values for expressing emotion.
As far as we know, this is the first system to build the deep learning model for the dimensional emotion model (VA model).
\item We build a new image emotion dataset through the crowdsourcing. The feed forward neural network is used to learn emotion features from this database. 
\item We propose a novel idea of using an pre-trained CNN to relate the main object and background of the image to the emotional feature. 
\item We show the effectiveness of an object to estimate an emotion of an image using correlations between emotional values of words and images. 
\end{itemize}

\section{Related Work}

Emotion of an image can be evoked by various factors. To figure out significant features for the emotion prediction problem, many researchers have considered various types of the features from color statistics to the art and the psychological features.
Machajdik et al.~\cite{machajdik2010affective} introduced an affective image classification system using psychology and art theory based features such as Itten's color contrast and rule of thirds.
Zhao et al.~\cite{zhao2014exploring} proposed to extract principles of art features for emotion classification. 
Similarly, Lu et al.~\cite{lu2012shape} computed the shape based features in natural images.
As a high level concept for representing the sentiment of an image, adjective-noun pairs were introduced by Borth et al. in~\cite{borth2013large}.

Despite the rise of deep learning studies, relatively few studies have attempted to address the emotion prediction of images using the deep network.
With the data set from Borth et al.~\cite{borth2013large}, Chen et al.~\cite{chen2014deepsentibank} classified the adjective-noun pairs using the CNN and achieved better accurate classification performance than Borth et al.~\cite{borth2013large}.  

Several studies utilized the pre-trained model for the image classification and transferred the learned parameter. 
By changing the number of outputs to be the same as the number of labels of their dataset, the classifier can be trained. 
For example, for a binary classification with a positive and a negative label, the number of output would be two. 
Some researchers~\cite{peng2015mixed,you2016building,xu2014visual, you2015robust, campos2015diving} used AlexNet's structure~\cite{krizhevsky2012imagenet} with pre-trained weight to train their sentiment prediction frameworks with different output numbers. 
Binary classification which gives either a positive or a negative label was considered in~\cite{xu2014visual,campos2017pixels,islam2016visual,you2015robust}.
Peng et al.~\cite{peng2015mixed} and You et al.~\cite{you2016building} trained the classifiers for seven and eight classes, respectively.
In most studies, an emotion classification model was created through transfer learning using a pre-trained model for image classification.
In general, fine-tuned models record better emotion classification performance compared to previous studies with shallow networks.
In this paper, we use a feed forward neural network using low level and high-level features rather than transfer learning method.
Our method is compared with transfer learning method by using same training and test dataset.

Compared to the previous CNN based emotion recognition systems which are limited to output only a few number of emotional states, we propose to use the valence-arousal model to represent emotions.
By using these two parameters that lie on a continuous space, we can represent emotions much better than the previous works.
To enable the learning of the VA values using the CNN, we build a large set of data with various emotions and obtain the V-A labels through the crowdsourcing survey.

\begin{figure}[!]
\centering
    \includegraphics[width=3.4in]{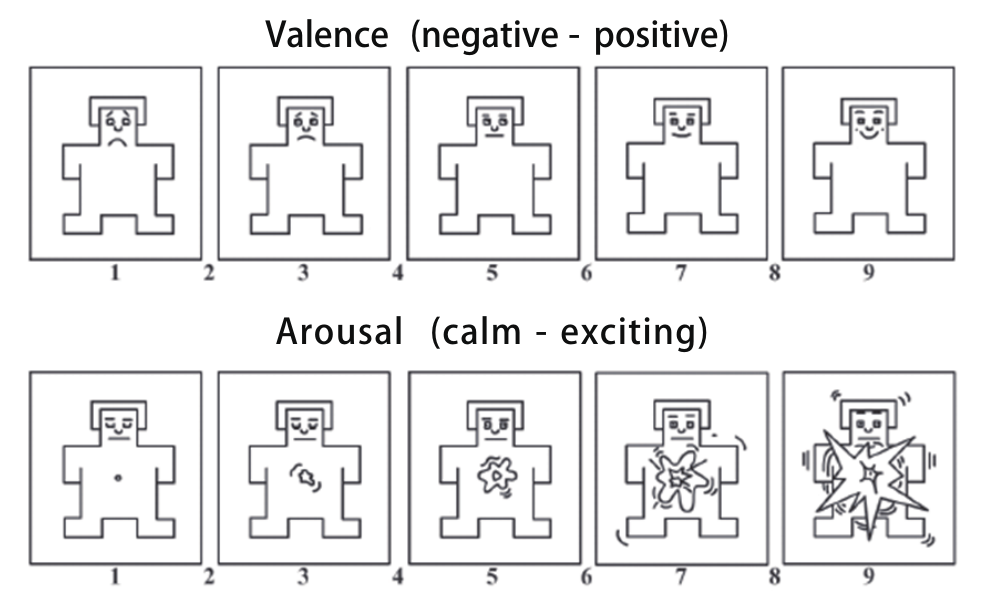}
\caption{Image representation of the Self Assessment Manikin (SAM)~\cite{bradley1994measuring} used for V-A value assignment in the user study.}
\label{fig:sam}
\end{figure}

\section{Image Emotion Dataset}
In order to build the dataset with emotion values (valence and arousal), we collected a large set of images from two resources. 
We first searched for the emotional images using the 22 keywords from~\cite{russell1980circumplex} in Flickr~\cite{flickr}.
Those words include what we call basic emotions (happy, angry, afraid, sad),
as well as less prototypical emotions (gloomy, bored) and affective states (sleepy, serene).

Using the keywords, we first collected over twenty thousand images. 
Since the goal is to collect emotional images, we manually eliminated non-emotional images from the collection.
Each candidate image in the dataset was assessed by three human subjects, and only the images that were determined to be emotional by more than two subjects were included in the database. 
The words used for the search are listed along with the number of images in the dataset: afraid (113), alarmed (14), angry (217), annoyed (179), distressed (150), frustrated (77), tense (8), aroused (4), delighted (28), excited (734), glad (315), happy (102), astonished (13), at ease (16), content (656), satisfied (33), serene (1917), pleased (42), depressed (179), bored (312), tired (942), gloomy(793). As a result, the total number of image is 6844. 

Second, we collected more images from~\cite{you2016building} which include eight types of emotion categories (Amusement, Awe, Contentment, Excitement, Anger, Disgust, Fear, Sad). 
We took 3,236 images out of 23,308 images.
An even number of images from eight classes have been selected and included in our database.
As a result, a total of 10766 images is acquired in our dataset. 

Next, we use the AMT to assign V-A emotion values for all images in our database.
Given an image, a worker rates the emotion values for each image using the representation of the V-A scale, Self-Assessment Manikin (SAM)~\cite{bradley1994measuring} (Figure~\ref{fig:sam}). 
In each question, two images are shown to the worker. One is the image whose emotion value is to be measured, and the other is the image of the previous question with emotion value recorded by the worker. 
By showing the previous image, the worker can measure a value for the given image compared to the value selected in the previous problem. It also alleviates the difficulty of having to choose absolute numbers.
Each image is represented to the worker randomly and is evaluated only once by the same worker. 
To ensure the quality of the answers, we only allow the workers with 95\% approval rate to participate in our Human Intelligence Task (HIT). 
We also limit the number of questions for each worker to 200 to keep them focused throughout the test.

\begin{figure}[!]
\centering
    \includegraphics[width=3.4in]{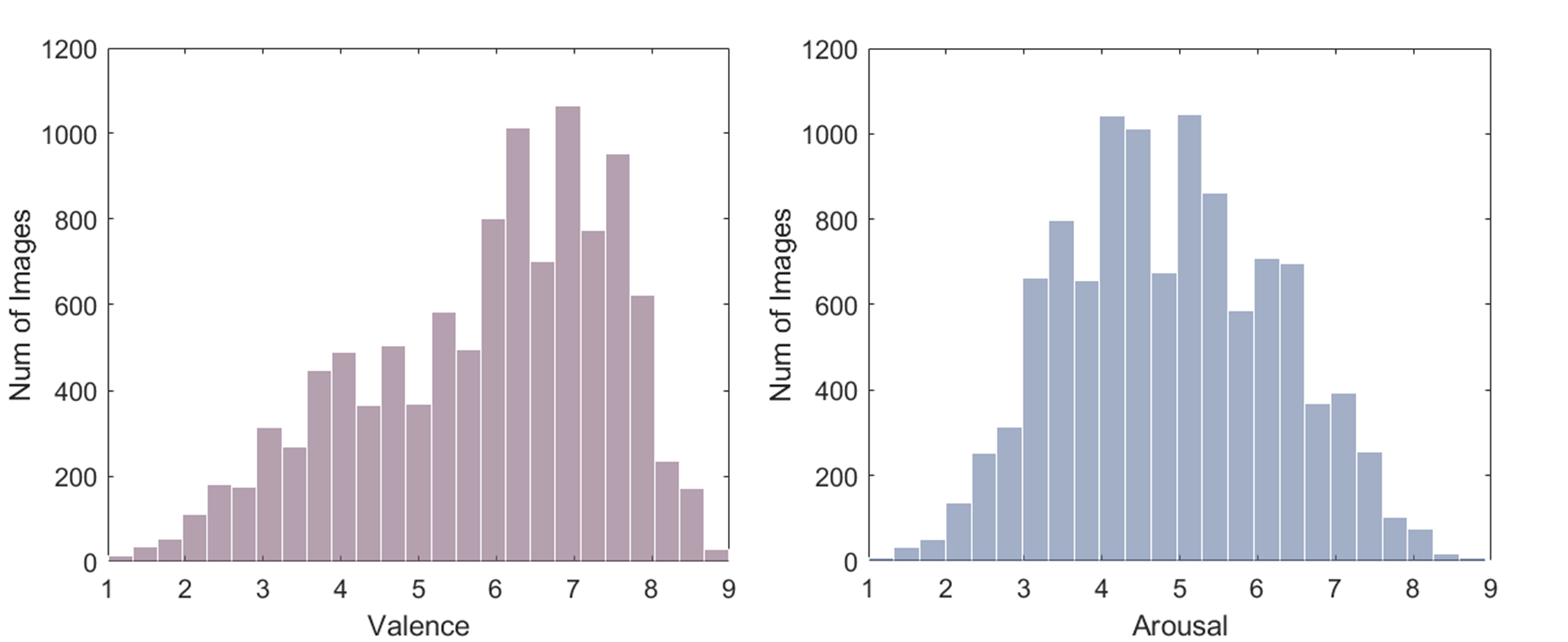}
\caption{The histogram of valence and arousal value from the collected images.}
\label{fig:avdist}
\end{figure}

\begin{figure*}[!]
\centering
    \includegraphics[width=6.5in]{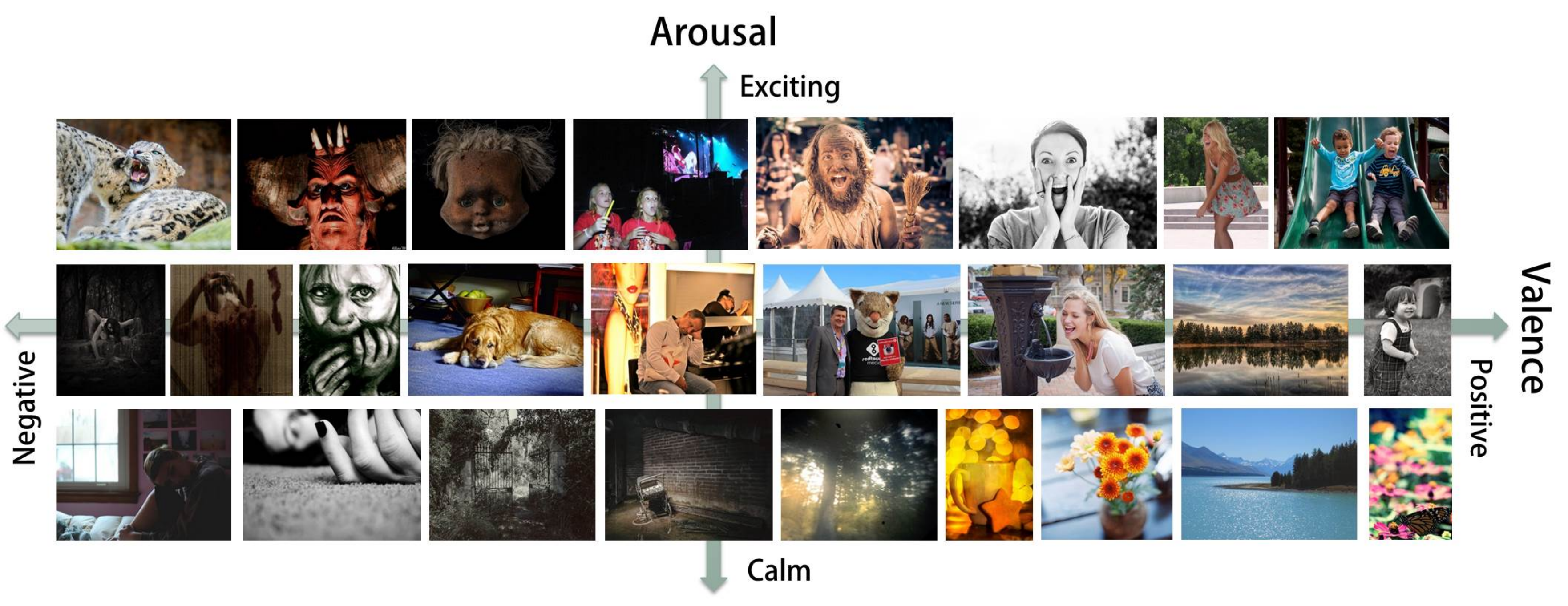}
\caption{Example images in our database. From left to right side, the valence value of the image increase.
  The images on the left/right have negative/positive emotion.
    From bottom to top, the arousal value of the image increase.
  The images on the bottom/top have the calm/exciting emotion.}
\label{fig:dataset}
\end{figure*}

A total of 1,339 workers were recruited to assign the emotion values of the 10,766 images in the dataset. 
The evaluation time per image varied from 10 seconds to a minute depending on the worker. 
Each image is evaluated on at least five workers, and the average of the acquired values is assigned to the emotion value of the image.
Figure~\ref{fig:avdist} shows the total distribution of the V-A values of the images. 
In the case of Valence, it can be seen that there are relatively more positive images than negative images.
In common sense, people usually share the positive images, rather than negative images, and this tendency is well represented in this distribution.
In the case of Arousal, various emotion values were obtained except for the extreme values.
Figure~\ref{fig:dataset} shows some of the images in our database.

\begin{figure}[!]
\centering
    \includegraphics[width=3.5in]{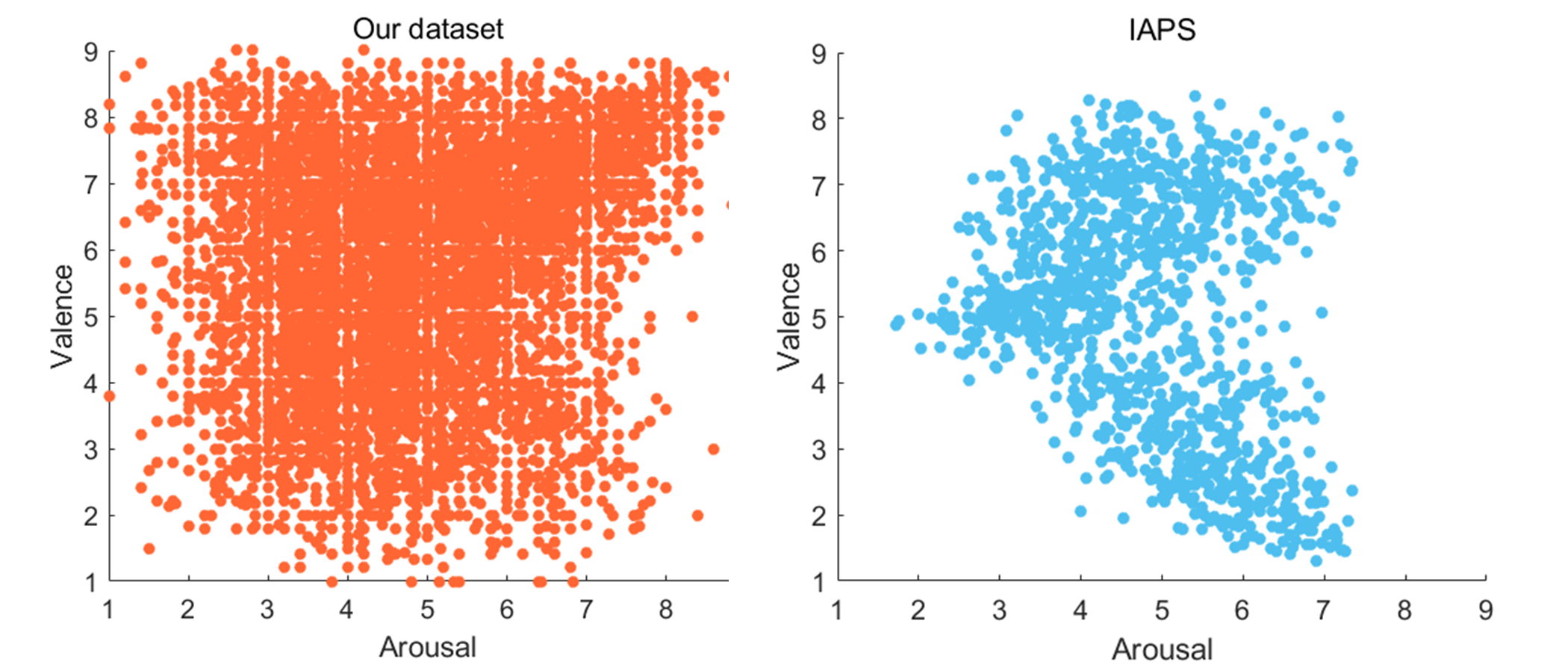}
\caption{Emotion distribution of our database and the IAPS~\cite{lang2008international} in V-A emotion space.}
\label{fig:emodist}
\end{figure}

We compare the emotion distribution of the dataset we collected with the distribution of IAPS dataset which is an emotion dataset introduced in~\cite{lang2008international}. 
The IAPS consists of
 of 1182 images, each of which contains valence, arousal and dominance values.
Although the number of images is different, we can see that our data is spread more than the IAPS data distribution in the V-A emotion space (Figure~\ref{fig:emodist}).

\begin{figure}[!]
\centering
    \includegraphics[width=3.4in]{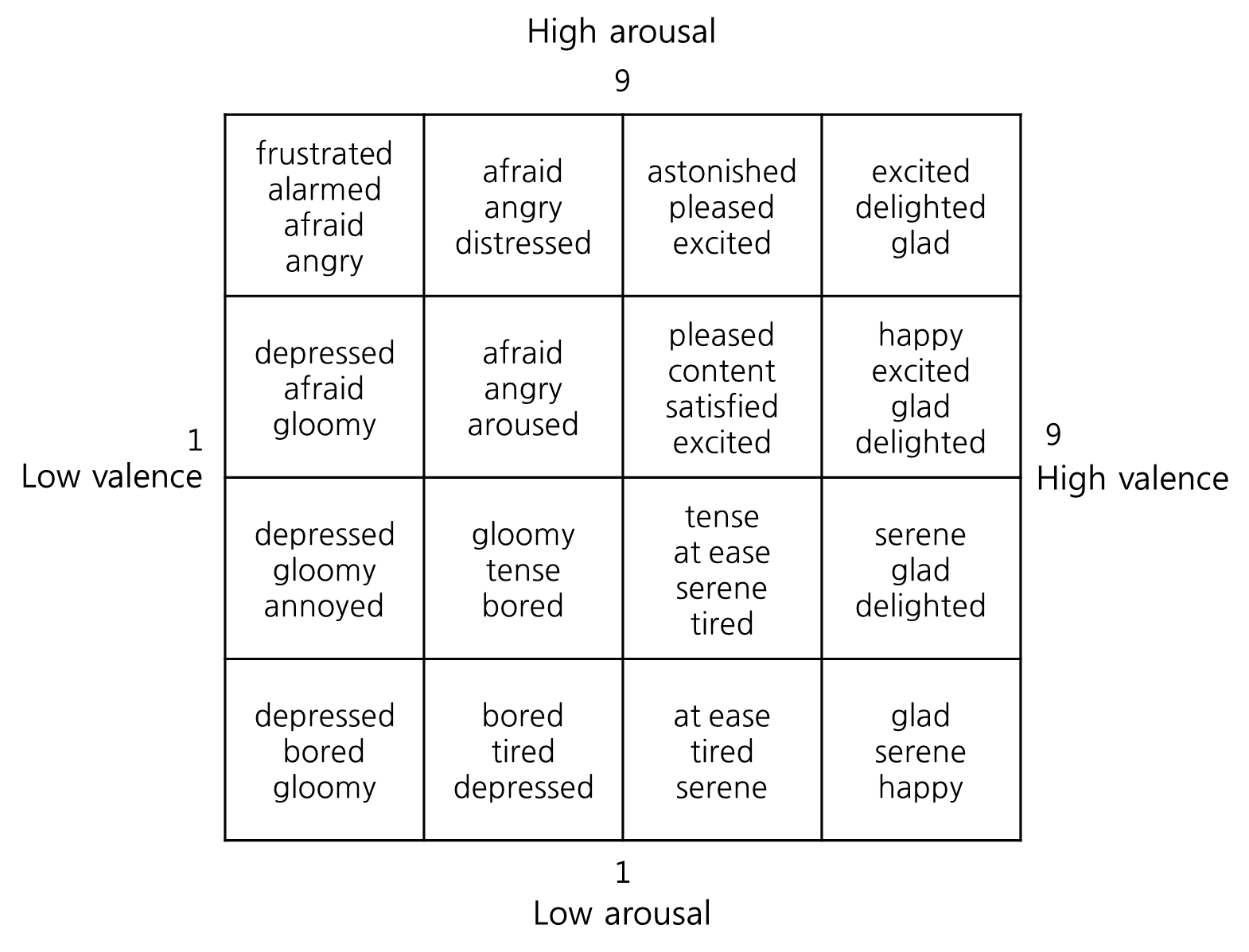}
\caption{The most used words in each section.}
\label{fig:worddist}
\end{figure}

We provide additional analysis to validate our new dataset. 
We split Valence-Arousal space into 4 by 4 sections: 
Most negative (from 1 to 3), negative (from 3 to 5), positive (from 5 to 7), and most positive (from 7 to 9). 
Most calm (from 1 to 3), calm (from 3 to 5), exciting (from 5 to 7), and most exciting (from 7 to 9). 
We then place the images in the emotion space according to the obtained emotion values and extract the most used words (using the image tags) in each section. 
The results are shown in Figure~\ref{fig:worddist}. 
In the negative section of V-A space, most words have low valence values such as afraid, angry, annoyed, depressed and gloomy. 
On the other hand, the words such as content, serene, excited, and glad are included in the positive section. 
In the case of the arousal, the lowest arousal side (from 1 to 3) consists of the words such as serene, tired and gloomy and the highest arousal side (from 7 to 9) include the words such as excited, angry and distressed. 
This analysis indicates that the use of words for image collection and user study is appropriate for obtaining as diverse and well-distributed emotion values as possible in space. 
The new image emotion dataset can be used for the emotion recognition in two ways.
A classifier can be trained to output either the continuous V-A values or discrete categories of emotions using the words in Figure~\ref{fig:worddist}.

\section{Features}
Now, we extract a various type of features for emotion prediction including color, local, object, and semantic features. 

\subsection{Color features}
Color is the most basic and powerful element to express emotions and can be effectively used by artists to induce emotional effects. 
Many studies have been conducted to change the color of an image as a means to change the emotion of the image~\cite{reinhard2001color,csurka2010learning,he2014image,kim2016image}.
Color is not an element that can directly resolve an affective-gap because it can be viewed as a low-dimensional feature, but color is still a crucial factor in emotion recognition.
We extract the mean values of RGB and HSV color space as the basic color characteristics.
We also calculate the HSV histogram and extract the label number and values of the bin with the largest value in the histogram.
A concept similar to a color histogram, it calculates how much of the 11 basic colors exist in the image~\cite{van2009learning}.
According to~\cite{osgood1952nature}, saturation and brightness can have direct in on pleasure, arousal, and dominance. Using the saturation and brightness values, Valdez and Mehrabian~\cite{valdez1994effects} introduced formulas to measure the value of pleasure, arousal, and dominance through experiments.
The formula for computing the values is as follows: 
\begin{align} 
Pleasure  = 0.69 Y + 0.22 S \\
Arousal   = 0.31 Y + 0.60 S \\
Dominance = 0.76 Y + 0.32 S
\end{align}
We measure the values for the three elements from the image and use them as features.

\begin{figure}[!]
\centering
    \includegraphics[width=3.2in]{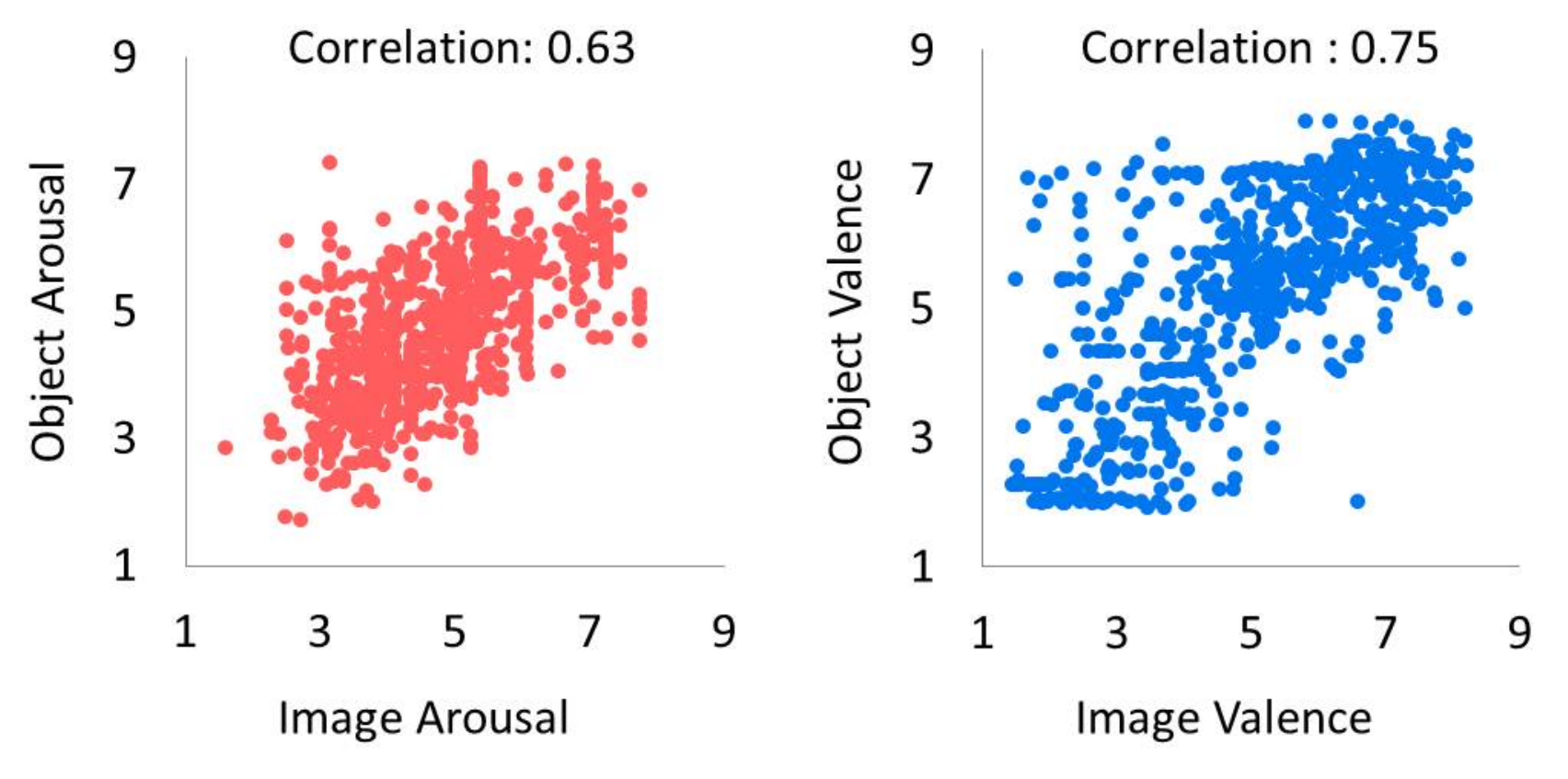}
\caption{Correlation between image emotion and object emotion. Image emotion values are taken from the IAPS data \cite{lang2008international} set and object emotion values are taken from the word emotion dictionary~\cite{warriner2013norms}.}
\label{fig:corr}
\end{figure}

\begin{figure*}[!]
\centering
    \includegraphics[width=6.7in]{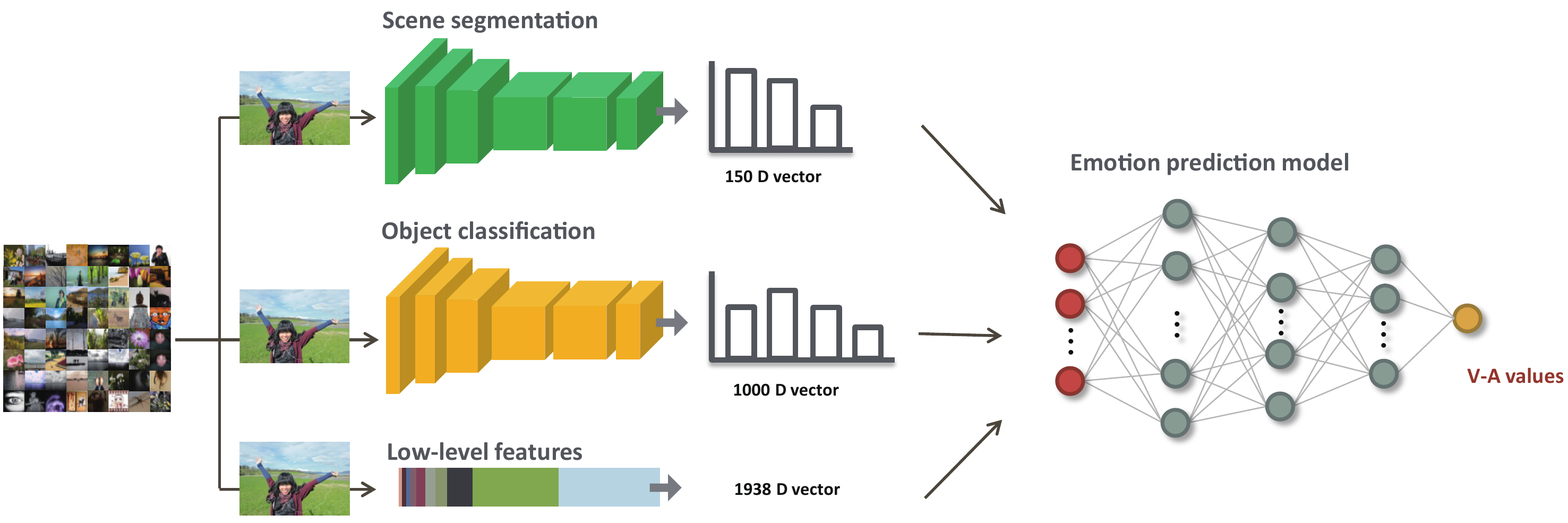}
\caption{The overall architecture of the proposed emotion prediction model.}
\label{fig:overview}
\end{figure*}

\subsection{Local features}
We exploit two kinds of local features used in the~\cite{borth2013large}. We use a 512-dimensional GIST descriptor that is effective in detecting scenes and a 59-dimensional local binary pattern (LBP) descriptor that is effective in detecting textures.

\subsection{Object features}
Identifying the emotion of the image from low-level features, such as color statistics and texture-related features, is difficult for a human subject.
Many researchers stated the need for high level of features on the affective level and designed the various types of features for emotion prediction. 
In our study, we assume that the object is one of the most important factors contributing to the emotion of an image. 
We conducted an experiment to prove that the object in images has relevance to the emotion elicited from the images. 
Each image in the IAPS dataset, an emotion dataset introduced in~\cite{lang2008international}, includes a tag representing the primary object (e.g., baby, snake, and shark) and the V-A emotion values. 
In order to replace the object tag with the emotion level representation, we adopt the word emotion dictionary~\cite{warriner2013norms} with the valence and the arousal values for each word.
We convert each tag to V-A values by searching for the tag in the word emotion dictionary. 
By measuring the Pearson correlation coefficient, we can easily understand the relationship between the object and the image emotion. 
Figure~\ref{fig:corr} shows the results of the experiment. As you can see, the correlation between the object emotion and the image emotion is significantly high, which means the object affects the emotion of the image. 
Especially, the correlation of the valence is higher than the arousal, 
which means that valence is more likely to be affected by the object than the arousal.

Based on this observation, we add object-based features to our system in predicting emotions. 
In recent years, many studies have used CNN models of various structures using ImageNet dataset~\cite{deng2009imagenet} for the image classification. 
We use three of the most popular models (AlexNet, VGG16, and ResNet) to extract object features from our image datasets and experiment with the effects of features extracted from each model on emotion prediction results. 
Our object feature is the result of the final output layer, and it represents the probabilities of 1000 object categories. 

\subsection{Sematic features}
As a high-level feature with a similar concept to object, we consider another semantic information that can describe the background of the image. 
It is important what the object is in the image, but what the background is made up of is as important. For example, if the main object of an image is a person, the emotion may be different depending on whether the background is a city with a lot of buildings or nature such as a mountain or a sea. Also, the ratio of sky, sea, or buildings in the background can also affect the emotions. To use the semantic information of the background as a feature, we perform scene parsing on all images.
Wu et al.~\cite{wu2016wider} proposed a semantic segmentation method based on a deep network, which classifies each pixel of an image into one of 150 semantic categories. Given a semantic map, we find out which of the 150 semantic categories each pixel in the image belongs to. As a result, a 150-dimensional vector is obtained and used as one of the input features.

These object and semantic-based high level features are combined with low level features such as color and local features to learn our networks for emotion prediction.

\section{Learning emotion model}
In this section, we introduce the details of our emotion prediction framework. 
The overall architecture of our framework is shown in Figure~\ref{fig:overview}. 
Our model is a fully connected feedforward neural network. 
In general, a neural network consists of an input layer, an output layer, and one or more hidden layers. 
Normally, when there are two or more hidden layers, the network is called deep network.
Each layer is made up of multiple neurons, and the edges that connect neurons between adjacent layers have weights.
The values of neurons (except the neurons in input layer) and weights are trained during a training phase.

Our network $F(X,\theta)$ including an input layer, three hidden layers and an output layer is as follows: 
\begin{equation}
F(X,\theta)=f^4 \circ g^3 \circ f^3 \circ g^2 \circ f^2 \circ g^1 \circ f^1(X), 
\end{equation}
where $X$ is input feature vector, $\theta$ is a set of weights including weights $w$ and bias $b$, and $f^4$ produces the final output of our neural network (Valence and Arousal value). 

Specifically, given an input vector $X^l={[x_i^l,...,x_n^l]}^T$ in layer $l$, the preactivation value $p_j$ for the neuron $j$ of layer $l+1$ is obtained through function $f^{l+1}$: 
\begin{equation}
p_j^{l+1} = f^{l+1}(x^l) =\sum_{ i=1 }^{ n }w^{l+1}_{ij}x_i^l+b_j^{l+1} , 
\end{equation}
where $w_{ij}^{l+1}$ is the connection weight connecting $x_i$ in layer $l$ to neuron $j$ in layer $l+1$, $b_j^{l+1}$ is the bias of neural j in layer $l+1$, and $n$ is the number of the neuron in layer $l$.
Then, the output value $x_j$ of the layer $l$ (also input vector for layer $l+1$) is obtained through function $g^l$ which is a nonlinear activation function in layer $l+1$, 
\begin{equation}
x_j^{l+1} =g^{l+1}(p_j^{l+1}).
\end{equation}
Note that, rectified linear unit(ReLU)~\cite{krizhevsky2012imagenet}, max(0, x), is used as the non-linear activation function throughout the all network.

The number of neurons in each layer is given in the Table~\ref{table:structure}.
We set the loss function of our network as $L=\sum_{ k=1 }^{ K }{(f_k^{4} - T_k)^2}$,
where the $f_k^{4}$ and $T_k$ are the output value predicted by our model and the ground truth emotion value of given image $I$ and $K$ is the number of images.
In training phase, network weights $\theta$ are updated by backpropagating the gradients through all layers.  
By minimizing the cost of the loss function, we can optimize the weights of our network. 
We set learning rate to 0.0001 and the network is trained by using the stocahstic gradient descent (SGD) optimization method with momentum of 0.9.  
We set the batch size to 1,000 and train our model until the error no longer diminishes.
All experiments are implemented by using the open source deep learning framework Tensorflow~\cite{abadi2016tensorflow}.

\begin{table}[!]
\centering
\caption{The number of neurons in each layer.}
\label{table:structure}
\begin{tabular}{|c|c|c|c|c|c|}
\hline
 & Input & H1 & H2 & H3 & Output \\ \hline
Num of neurons & 1588 & 3000 & 1000 & 1000 & 1 \\ \hline
\end{tabular}
\end{table}

\section{Experiment}

\subsection{Model performance}
We first evaluate the performance of our model.
The entire dataset is divided into five groups, four groups are used for the training phase, and the remaining one group is used for the test phase. 
Each group is used as a test phase once.
In each group, the number of training images is 8600, and the number of test images is 2166.
All groups are learned using the same structure.

The results are presented in Table~\ref{table:modelresult}. 
The number in the first row represents the group number.
Column g1 shows the training and test error when using group 2 to 5 as training data and group 1 as test data. The dimension of input feature is 3088 described in Tabel~\ref{table:structure}, and the output is valence or arousal value. 
To compare the performance of the object features obtained from the three CNNs, we build various models by combining object features with other features and compare their performance.
Note that the `A', `V', and `R' in the third column represent the AlexNet, VGG16 and ResNet, respectively. The number of each bin represent the mean square error between ground truth emotion value and predicted emotion value by each model. 
We also extracted category-level features from~\cite{yu2013designing}, not from CNN-based model, and included them as input features in emotion prediction. Borth et al.~\cite{borth2013large} also used this feature to predict the sentiment of images.
Note that, the dimension of this feature is 2000, and the number of input feature neurons in our model is changed to 2588.
The row with 'O' represents the prediction result.
The results show that the features from VGG16 achieved the best performance in both the 'Valence' and 'Arousal' models (valence: 1.64, arousal: 1.47).

\begin{table}[!]
\centering
\caption{Results of 5-fold validation experiments}
\label{table:modelresult}
\begin{tabular}{|c|c|c|c|c|c|c|c|c|}
\hline
 &  &  & g1 & g2 & g3 & g4 & g5 & Avg. \\ \hline
\multirow{8}{*}{Valence} & \multirow{4}{*}{Training} & A & 1.37 & 1.37 & 1.35 & 1.37 & 1.38 & 1.37 \\ \cline{3-9} 
 &  & V & 1.31 & 1.31 & 1.30 & 1.30 & 1.33 & 1.31 \\ \cline{3-9} 
 &  & R & 1.29 & 1.30 & 1.28 & 1.31 & 1.30 & 1.30 \\ \cline{3-9} 
 &  & \multicolumn{1}{l|}{O} & \multicolumn{1}{l|}{1.47} & \multicolumn{1}{l|}{1.47} & \multicolumn{1}{l|}{1.46} & \multicolumn{1}{l|}{1.48} & \multicolumn{1}{l|}{1.48} & \multicolumn{1}{l|}{1.47} \\ \cline{2-9} 
 & \multirow{4}{*}{Test} & A & 1.72 & 1.68 & 1.67 & 1.69 & 1.61 & 1.67 \\ \cline{3-9} 
 &  & V & 1.68 & 1.66 & 1.63 & 1.63 & 1.59 & \textbf{1.64} \\ \cline{3-9} 
 &  & R & 1.70 & 1.66 & 1.65 & 1.62 & 1.60 & 1.65 \\ \cline{3-9} 
 &  & \multicolumn{1}{l|}{O} & \multicolumn{1}{l|}{1.80} & \multicolumn{1}{l|}{1.81} & \multicolumn{1}{l|}{1.73} & \multicolumn{1}{l|}{1.73} & \multicolumn{1}{l|}{1.69} & \multicolumn{1}{l|}{1.75} \\ \hline
\multirow{8}{*}{Arousal} & \multirow{4}{*}{Training} & A & 1.22 & 1.24 & 1.20 & 1.21 & 1.23 & 1.22 \\ \cline{3-9} 
 &  & V & 1.16 & 1.21 & 1.17 & 1.16 & 1.16 & 1.17 \\ \cline{3-9} 
 &  & R & 1.18 & 1.22 & 1.18 & 1.18 & 1.18 & 1.19 \\ \cline{3-9} 
 &  & \multicolumn{1}{l|}{O} & \multicolumn{1}{l|}{1.26} & \multicolumn{1}{l|}{1.30} & \multicolumn{1}{l|}{1.25} & \multicolumn{1}{l|}{1.27} & \multicolumn{1}{l|}{1.29} & \multicolumn{1}{l|}{1.27} \\ \cline{2-9} 
 & \multirow{4}{*}{Test} & A & 1.50 & 1.52 & 1.50 & 1.48 & 1.44 & 1.49 \\ \cline{3-9} 
 &  & V & 1.48 & 1.49 & 1.46 & 1.48 & 1.44 & \textbf{1.47} \\ \cline{3-9} 
 &  & R & 1.49 & 1.49 & 1.47 & 1.47 & 1.45 & 1.48 \\ \cline{3-9} 
 &  & \multicolumn{1}{l|}{O} & \multicolumn{1}{l|}{1.56} & \multicolumn{1}{l|}{1.53} & \multicolumn{1}{l|}{1.54} & \multicolumn{1}{l|}{1.50} & \multicolumn{1}{l|}{1.48} & \multicolumn{1}{l|}{1.52} \\ \hline
\end{tabular}
\end{table}

\begin{figure*}
\center
\includegraphics[width=1.0\linewidth]{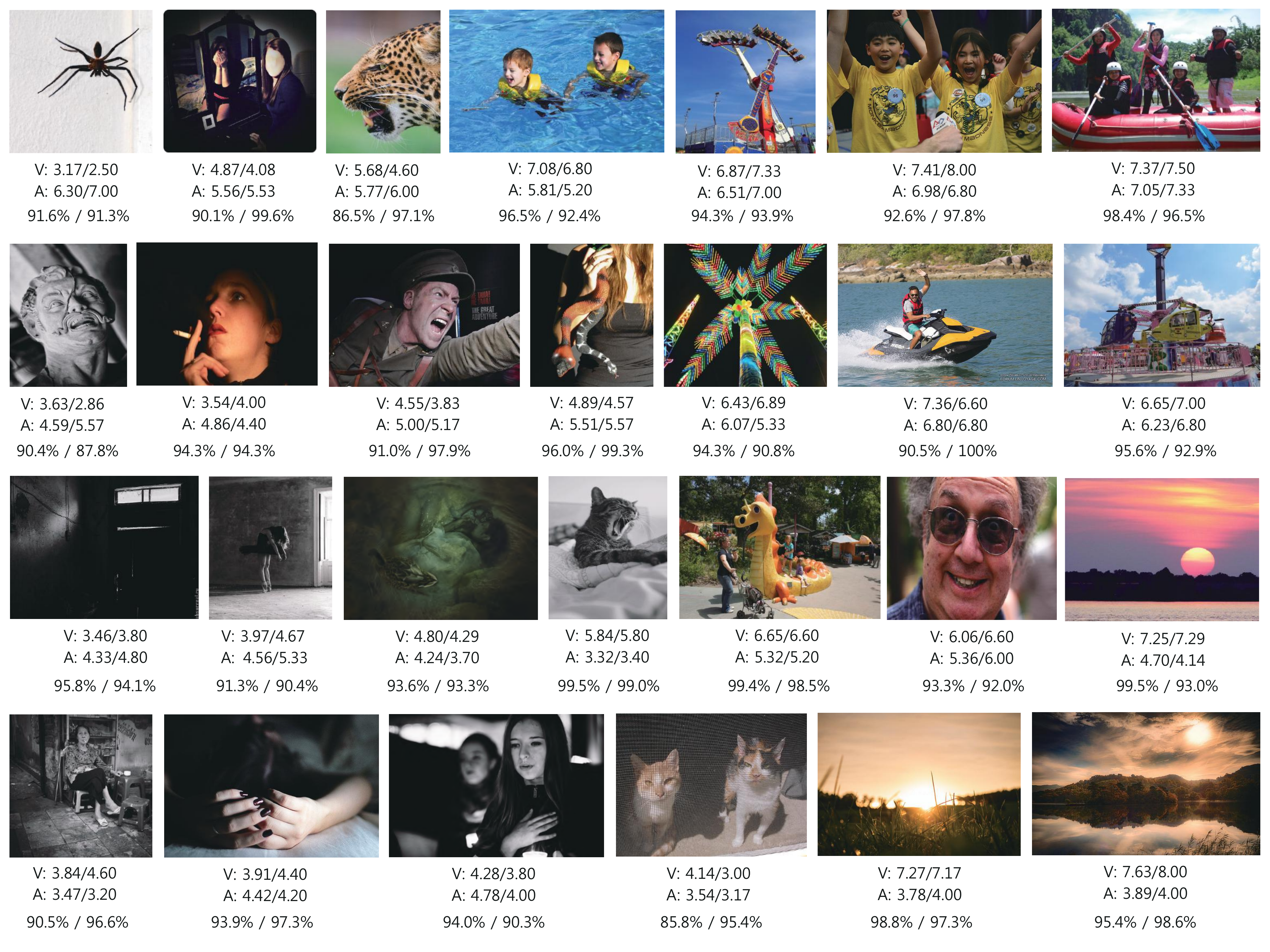} 
   \caption{Prediction results. The values below each image show the prediction results of FFNN and ground truth emotion values, respectively. The prediction accuracy results (V/A) are also shown. 
}   
\label{fig:result}
\end{figure*}

\begin{figure*}
\center
\includegraphics[width=0.98\linewidth]{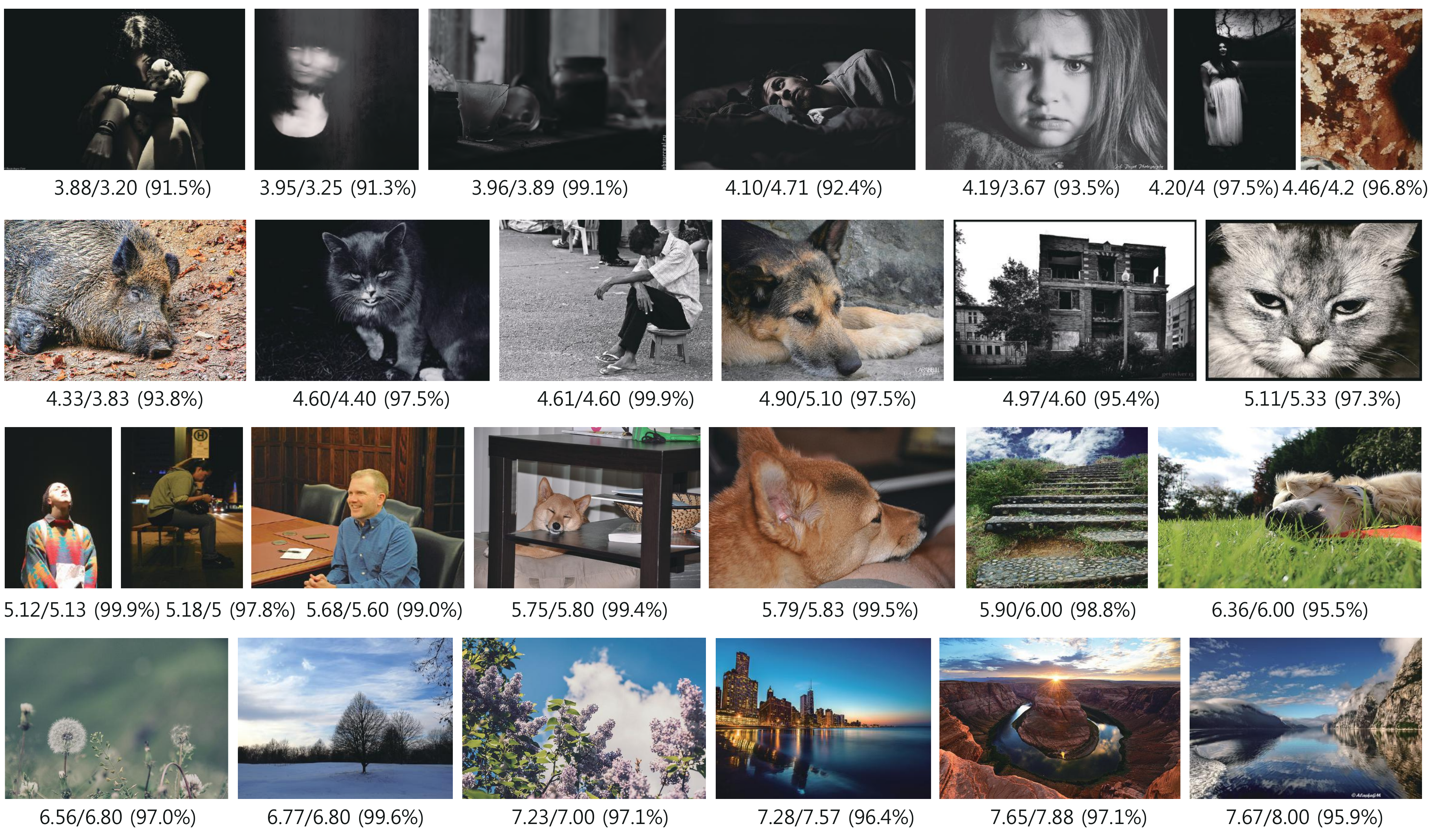} 
	\caption{Valence prediction results of the images with low arousal. The values below each image represent the predicted valence value and ground truth value (prediction/ground truth). The prediction accuracy result is also shown. From the top left to the bottom right, the value of valence increases. All images in this example have low arousal values.}

\label{fig:result_valence1}
\end{figure*}

\begin{figure*}
\center
\includegraphics[width=0.98\linewidth]{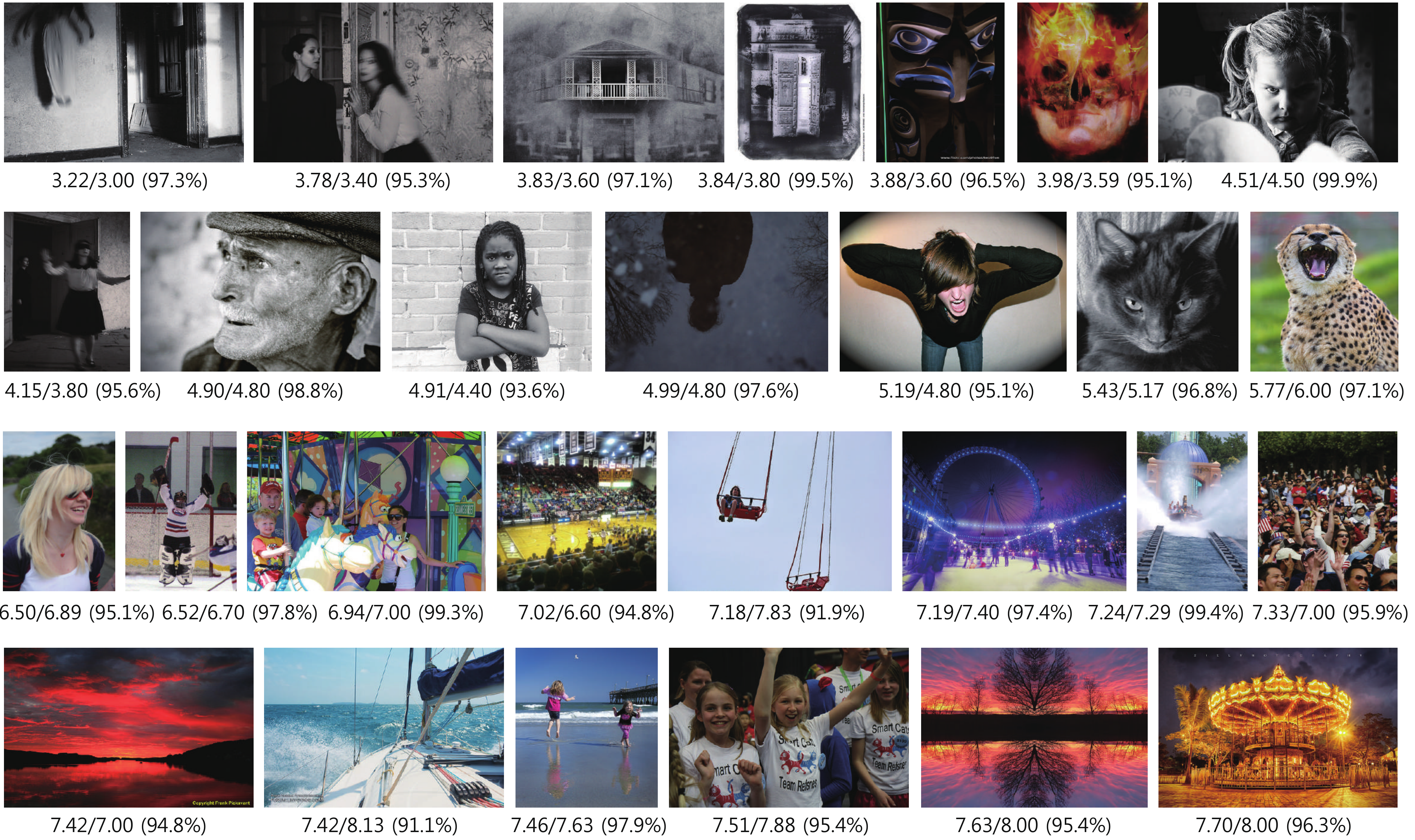} 
	\caption{Valence prediction results of the images with high arousal. The values below each image represent the predicted valence value and ground truth value (prediction/ground truth). The prediction accuracy result is also shown. From the top left to the bottom right, the value of valence increases. All images in this example have high arousal values.}

\label{fig:result_valence2}
\end{figure*}

\begin{figure*}
\center
\includegraphics[width=0.98\linewidth]{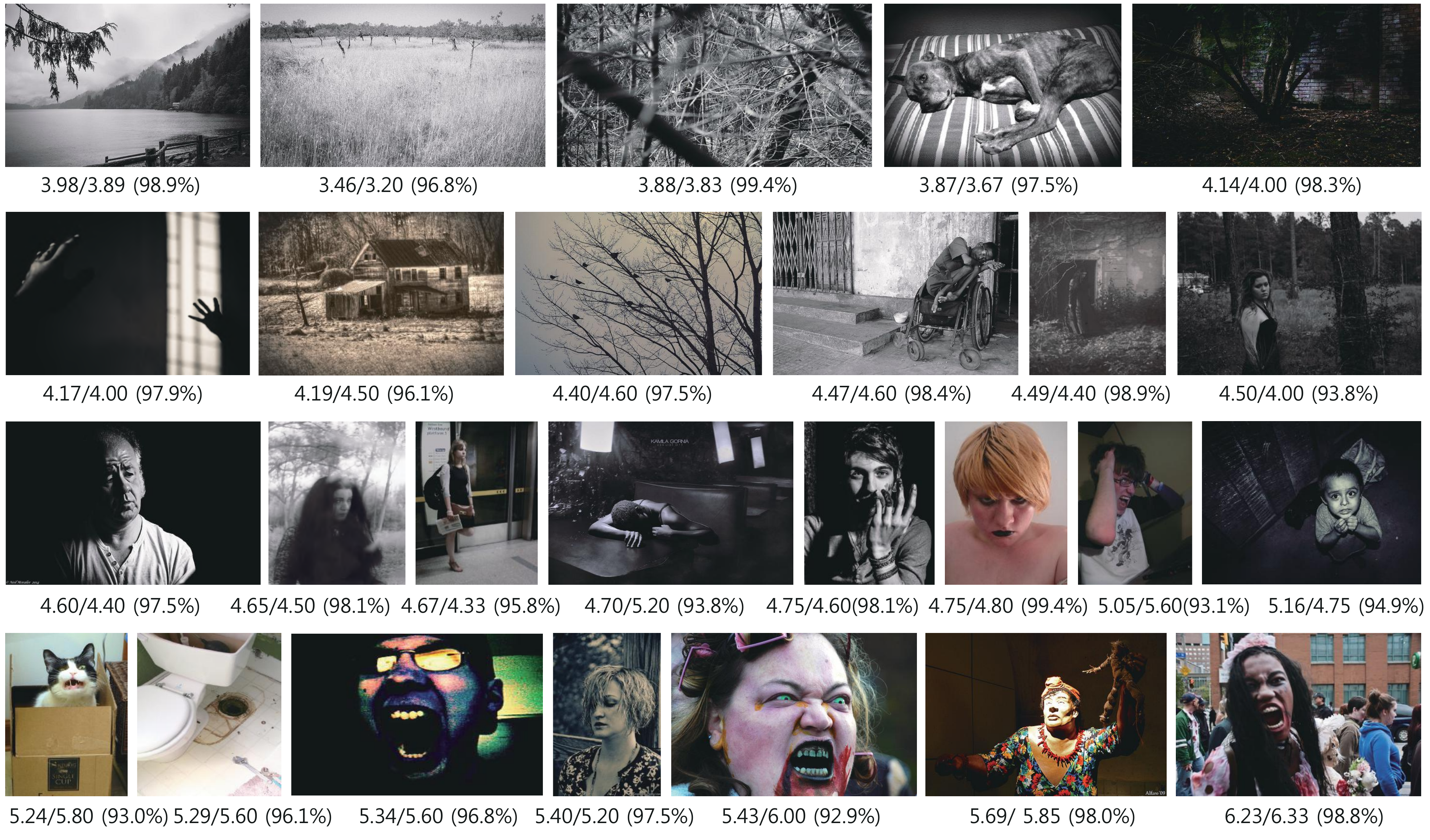} 
   \caption{Arousal prediction results of the images with low valence. The values below each image represent the predicted arousal value and ground truth value (prediction/ground truth). The prediction accuracy result is also shown. From the top left to the bottom right, the value of arousal increases. All images in this example have low valence values. 
}   
\label{fig:result_arousal1}
\end{figure*}

\begin{figure*}
\center
\includegraphics[width=0.98\linewidth]{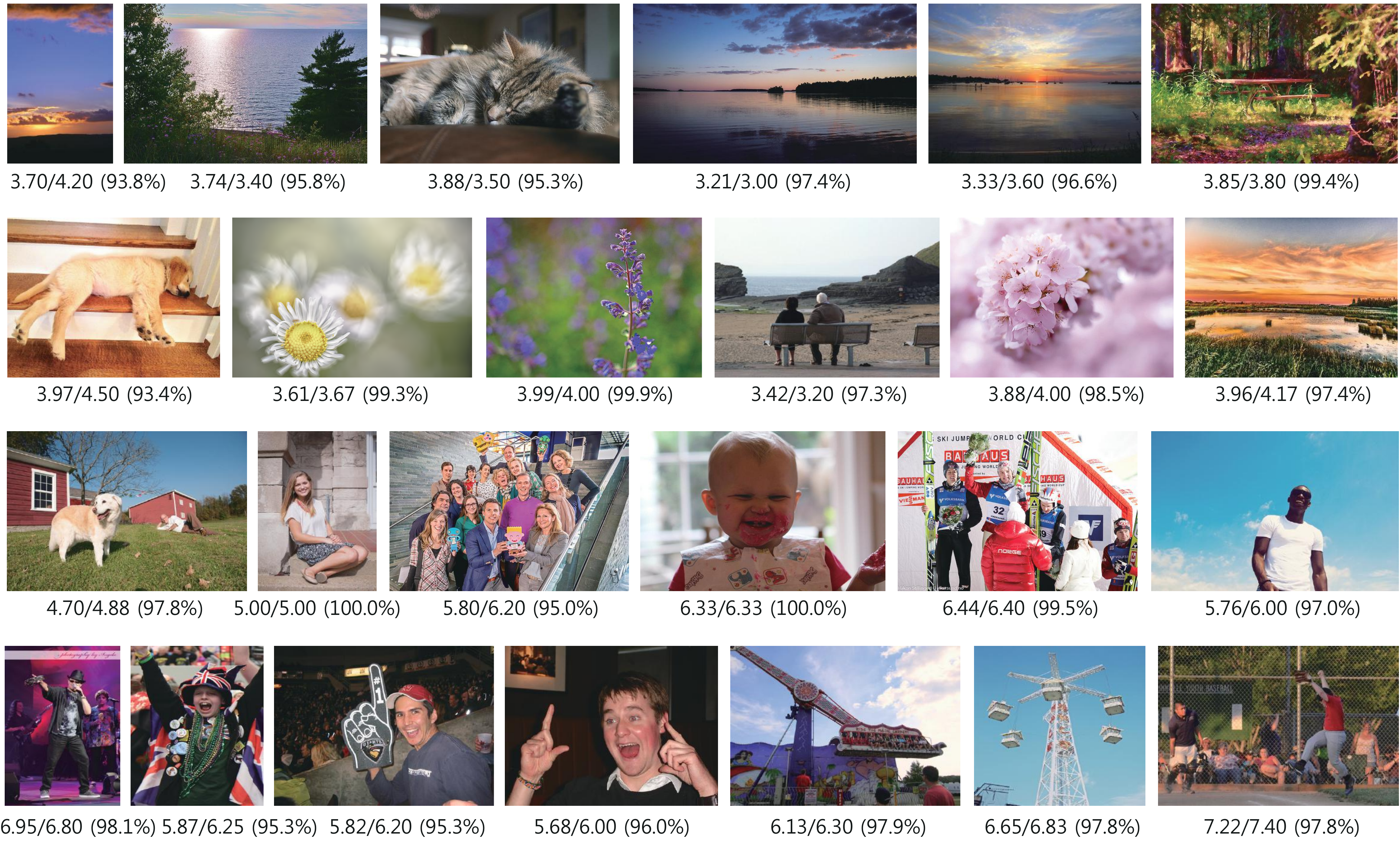} 
   \caption{Arousal prediction results of the images with high valence. The values below each image represent the predicted arousal value and ground truth value (prediction/ground truth). The prediction accuracy result is also shown. From the top left to the bottom right, the value of arousal increases. All images in this example have high valence values. 
}   
\label{fig:result_arousal2}
\end{figure*}

\Cref{fig:result_valence1,fig:result_valence2,fig:result_arousal1,fig:result_arousal2,fig:result} show the qualitative analysis with emotion values and accuracy which is predicted by our model. 
In Figure~\ref{fig:result}, the images were placed so that the predicted values matches the emotion values in the VA space.
Figure~\ref{fig:result_valence1} and \ref{fig:result_valence2} shows the results of the valence model.  Figure~\ref{fig:result_arousal1} and \ref{fig:result_arousal2} shows the results of the arousal model.

\subsection{Feature performance}
We also investigate the effects of the various features we proposed.
First, we combine the color feature and the local features into a low level feature.
As a method for constructing a network model using each feature, we use the structure of our model and change the number of neurons. Our proposed model consists of 3 hidden layers with 3000, 1000, and 500 neurons.
In the model for feature learning, the number of nodes in each layer is based on the ratio of the number of nodes in two adjacent layers of our model (Table~\ref{table:featureresult} (bottom)).
As a result, the object feature among the three features showed the best result in emotion prediction. In Valence, the object feature extracted from VGG16 had the best result (mse:1.92). In arousal, the object feature extracted from AlexNet showed the best result (mse:1.61).
Semantic features also showed lower error than low-level features. Our model combining all the features resulted in the best prediction performance, which is the synergy effect of extracted features from various sources and deep neural network, which is a powerful expression power.

\begin{table}[h]
\centering
\caption{Performance of each feature}
\label{table:featureresult}
\begin{tabular}{ccccccc}
\hline
\multicolumn{1}{|c|}{\multirow{2}{*}{}} & \multicolumn{1}{c|}{\multirow{2}{*}{Low}} & \multicolumn{3}{c|}{Object} & \multicolumn{1}{c|}{\multirow{2}{*}{semantic}} & \multicolumn{1}{c|}{\multirow{2}{*}{All}} \\ \cline{3-5}
\multicolumn{1}{|c|}{} & \multicolumn{1}{c|}{} & \multicolumn{1}{c|}{Alexnet} & \multicolumn{1}{c|}{VGG16} & \multicolumn{1}{c|}{ResNet} & \multicolumn{1}{c|}{} & \multicolumn{1}{c|}{} \\ \hline
\multicolumn{1}{|c|}{Valence} & \multicolumn{1}{c|}{1.98} & \multicolumn{1}{c|}{1.93} & \multicolumn{1}{c|}{1.92} & \multicolumn{1}{c|}{1.98} & \multicolumn{1}{c|}{1.97} & \multicolumn{1}{c|}{1.64} \\ \hline
\multicolumn{1}{|c|}{Arousal} & \multicolumn{1}{c|}{1.66} & \multicolumn{1}{c|}{1.61} & \multicolumn{1}{c|}{1.62} & \multicolumn{1}{c|}{1.68} & \multicolumn{1}{c|}{1.64} & \multicolumn{1}{c|}{1.47} \\ \hline
\multicolumn{1}{l}{} & \multicolumn{1}{l}{} & \multicolumn{1}{l}{} & \multicolumn{1}{l}{} & \multicolumn{1}{l}{} & \multicolumn{1}{l}{} & \multicolumn{1}{l}{} \\ \hline
\multicolumn{1}{|c|}{input} & \multicolumn{1}{c|}{438} & \multicolumn{1}{c|}{1000} & \multicolumn{1}{c|}{1000} & \multicolumn{1}{c|}{1000} & \multicolumn{1}{c|}{150} & \multicolumn{1}{c|}{1588} \\ \hline
\multicolumn{1}{|c|}{h1} & \multicolumn{1}{c|}{900} & \multicolumn{3}{c|}{2000} & \multicolumn{1}{c|}{300} & \multicolumn{1}{c|}{3000} \\ \hline
\multicolumn{1}{|c|}{h2} & \multicolumn{1}{c|}{300} & \multicolumn{3}{c|}{700} & \multicolumn{1}{c|}{100} & \multicolumn{1}{c|}{1000} \\ \hline
\multicolumn{1}{|c|}{h3} & \multicolumn{1}{c|}{150} & \multicolumn{1}{c|}{350} & \multicolumn{1}{c|}{350} & \multicolumn{1}{c|}{350} & \multicolumn{1}{c|}{50} & \multicolumn{1}{c|}{500} \\ \hline
\multicolumn{1}{|c|}{output} & \multicolumn{6}{c|}{1} \\ \hline
\end{tabular}
\end{table}

\subsection{Comparison with CNN}
Some studies have learned emotion classification models using pre-trained weights learned for image classification~\cite{peng2015mixed,you2016building,xu2014visual, you2015robust, campos2015diving}.
We compare our emotion prediction model with CNN-based emotion prediction model generated by transfer learning.
Two CNN structures are used for comparison; AlexNet and Vgg19.
We first initialize the weights of AlexNet and VGG19 to the weights learned for the image classification.
Except for the final output layer, the other convolution layer and the fully-connected layer use the existing model structure. Since the number of the output layer of CNN model based on ImageNet is 1000, we change the number of output layers to 1 for our purpose (Valence and Arousal).

Besides, various results can be obtained in transfer learning. AlexNet has five convolutional layers and three fully connected layers, and VGG19 has 16 convolutional layers and three fully connected layers. In transfer learning, we can determine which layer to freeze and which layer to train.
We experiment with two conditions. 
The first is that the convolutional layer is frozen, only the fully connected layer is learned (conv-frozen), and the second is learning all layers together (conv-train).
The learning environment for the CNN-based model is almost similar to that of our FFNN model except for the batch size and learning rate. 
The learning rates of both conv-frozen network and conv-train network are $0.5\times10^{-4}$, and the batch size of both CNN models is 50. 
When we train the CNN models, including our model, we use same training and test dataset.

The results are shown in the Table~\ref{table:transferresult}.
The second column shows the training range, conv-frozen means that only the fully connected layer has been learned, and conv-train means that all layers have been learned.
From the training range perspective, it can be seen that the error of the transfer learning of the entire network is smaller than that of the fully connected layer transfer learning. 
This result implies that the filter in low-level, as well as the filters in high-level, must be learned in order to achieve a better performance emotion prediction model. 
In other words, the convolutional layer and the fully connected layer must be learned together. 
On the model side, we can see that the performance of the model learned by using the structure of Vgg19 Network is better than AlexNet. However, the results of both models are much more error-prone than our proposed emotion based FFNN.
If CNN-based models have the same type of data with the same class, such as object detection or image classification, the learning is well done and the prediction performance is excellent. However, as mentioned earlier, images with different shapes can have the same emotions, and images with similar shapes can have different emotions.
We also conducted the test using other machine learning methods with same dataset. Linear regression and Support vector regression method were used. 
Compared to the CNN-based model, the performance of both models is better, but the results of our model still show the best performance (See Table~\ref{table:transferresult}). 

\begin{table}[]
\centering
\caption{Comparision with other learning method}
\label{table:transferresult}
\begin{tabular}{|c|c|c|c|c|c|c|}
\hline
Emotion                  & Train range & AlexNet & VGG19 & Linear                & SVR                   & Ours                  \\ \hline
\multirow{2}{*}{Valence} & conv-frozen & 2.76    & 2.64  & \multirow{2}{*}{2.42} & \multirow{2}{*}{1.75} & \multirow{2}{*}{1.64} \\ \cline{2-4}
                         & conv-train  & 2.64    & 2.60  &                       &                       &                       \\ \hline
\multirow{2}{*}{Arousal} & conv-frozen & 1.95    & 1.91  & \multirow{2}{*}{2.65} & \multirow{2}{*}{1.54} & \multirow{2}{*}{1.47} \\ \cline{2-4}
                         & conv-train  & 1.89    & 1.87  &                       &                       &                       \\ \hline   
\end{tabular}
\end{table}

\section{Conclusion}
In this paper, we presented a new emotion recognition system with a deep learning framework.
To reduce the affective gap, we designed and extracted objects and background semantic features as high-level features, and showed that these features are effective for emotion prediction.
Both high-level features and low-level features complement each other well, which leads to better emotion recognition performance.
As expected, the accuracy of the object recognition has an impact on the performance of the emotion prediction. The object features with incorrect recognition may lead to incorrect emotion prediction results. There is also a problem when the main object of the image is not included in the existing 1000 classes. 
However, with the rapid progress in the deep learning technology with large dataset, the accuracy as the number of classes will be increased, which in turn will also help our emotion recognition system.

\begin{figure}
\center
\includegraphics[width=1.0\linewidth]{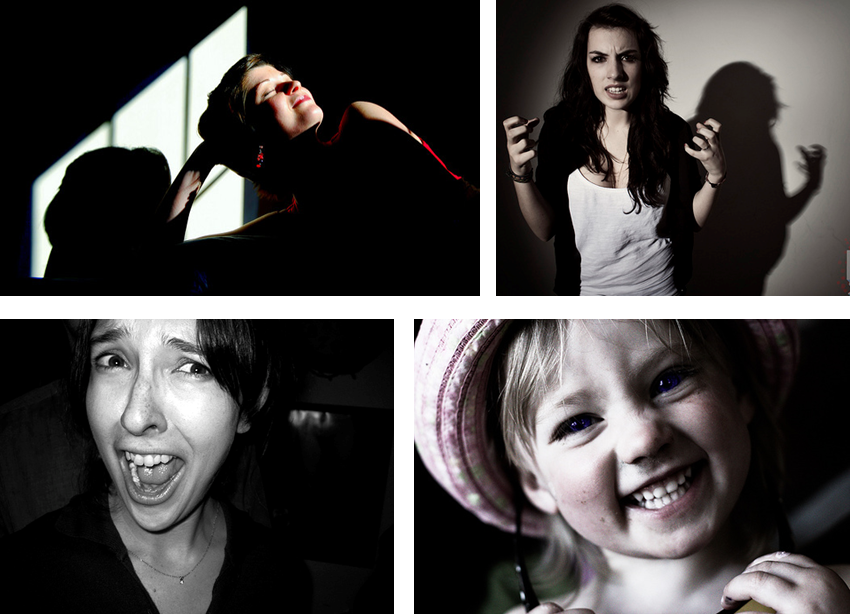} 
   \caption{Different emotions with different facial expressions.
}   
\label{fig:future}
\end{figure}

As an interesting future work, one can consider the presence of a person in the image and the facial expression.
A facial expression is one of the features that can greatly affect the emotion prediction. 
Even if the overall mood of the image is dark, smiling face can mitigate negative emotion a little (Figure~\ref{fig:future}).
In addition, when the face occupies most of the part in the photograph, facial expression and emotion are directly connected.
We will consider enhancing emotion recognition performance by adding a facial expression recognition framework.

Several studies have demonstrated that biometric data has a positive effect on emotion recognition~\cite{aftanas2004analysis,chanel2006emotion}.
We can also consider using biometric data or an observer's facial feature as additional features. 
However, in general, deep networks require thousands or tens of thousands of data, and collecting these biometric and facial data is not easy task. 
We will try to find the method to improve the performance of the model by considering a small amount of biometric data, which is left as another future work. 
  
We also built a database for the emotion estimation with the V-A model and will continue to collect more data. We expect our dataset will be widely used in the field of affective computing.

\ifCLASSOPTIONcaptionsoff
  \newpage
\fi



\bibliographystyle{IEEEtran}
\bibliography{IEEEabrv,emo_ref}
\end{document}